%% file: main.tex
\documentclass[twocolumn]{article}

\usepackage{arxiv}

\usepackage[utf8]{inputenc} 
\usepackage{hyperref}       
\usepackage{url}            
\usepackage{booktabs}       
\usepackage{amsfonts}       
\usepackage{nicefrac}       
\usepackage{microtype}      
\usepackage{graphicx}
\usepackage{doi}
\usepackage{textcomp}
\usepackage{multirow}
\usepackage[abs]{overpic}
\usepackage[dvipsnames]{xcolor}
\usepackage{graphicx}
\usepackage{float}
\usepackage{tikz}
\usepackage{amsmath}
\usepackage{pdfpages}
\usepackage{listings}
\usepackage{comment}
\usepackage{ftnright}
\usepackage{enumitem}
\usepackage{adjustbox}
\usepackage{subcaption}
\usepackage{mathtools}
\usepackage{pgfplots}

\usepackage[
backend=bibtex,
style=numeric,
citestyle=numeric
]{biblatex}
\setitemize{noitemsep,topsep=0pt,parsep=0pt,partopsep=0pt,leftmargin=*}
\setlist[enumerate]{noitemsep,topsep=0pt,parsep=0pt,partopsep=0pt,leftmargin=*}

\newcommand{\R}[1]{{\color{red}#1}}

\title{\textbf{VLRM: Vision-Language Models act as \\Reward Models for Image Captioning}}

\author{
	Dzabraev Maksim$^{1}$, Kunitsyn Alexander$^{1}$, Ivanyuta Andrey$^{1}$ \\
	$^1$Huawei Moscow Research Center \\
	dzabraev.maksim1@huawei.com,
	kunitsyn.alexander@huawei.com,
	ivanyuta.andrey@huawei.com
}

\renewcommand{\headeright}{}
\renewcommand{\undertitle}{}
\renewcommand{\shorttitle}{\textbf{VLRM: Vision-Language Models act as Reward Models for Image Captioning}}

\hypersetup{
pdftitle={VLRM - Vision-Language Models act as Reward Models for Image Captioning},
pdfauthor={M.Dzabraev, A.Kunitsyn, A.Ivanyuta},
pdfkeywords={image captioning, reinforcement learning, blip, clip},
}

\begin{document}
\arraycolsep=1.4pt
\setlength{\abovedisplayskip}{0pt}
\setlength{\belowdisplayskip}{0pt}
\setlength{\abovedisplayshortskip}{0pt}
\setlength{\belowdisplayshortskip}{0pt}

\twocolumn

\twocolumn[
  \begin{@twocolumnfalse}
    \maketitle
  \end{@twocolumnfalse}
]

\centerline{\bf Abstract}

In this work, we present an unsupervised method for enhancing an image captioning model (in our case, BLIP2) using reinforcement learning and vision-language models like CLIP and BLIP2-ITM as reward models.
The RL-tuned model is able to generate longer and more comprehensive descriptions.
Our model reaches impressive 0.90 R@1 CLIP Recall score on MS-COCO Carpathy Test Split.

Weights are available at \url{https://huggingface.co/sashakunitsyn/vlrm-blip2-opt-2.7b}.


\section{Introduction} \label{sec:introduction}
\input{sect_intro.tex}

\section{Related work}
\input{sect_related_work.tex}

\section{Method}
\input{sect_method.tex}

\section{Experiments}

\input{sect_exp.tex}

\section{Results}\label{sec:results}

\input{sect_results.tex}


\section{Conclusion}
\input{sect_conclusion.tex}

\input{sect_appendix.tex}

\clearpage

\printbibliography

\end{document}

%% file: sect_intro.tex
Recently developed image captioning models \cite{BLIP2, BLIP, coca} have demonstrated impressive performance.
However, they often lack fine-grained details in their captions, providing only the most essential and limited information about the scene.

The origin of the issue is in the training data used.
Both unsupervised \cite{laion400m, laion5b} and supervised \cite{mscoco, conceptualcaptions} captions tend to focus on the main object and action, often omitting specific details and rarely capturing all relevant information.
This is the reason why image captioning datasets typically provide several different captions for each image, aiming to ensure comprehensive coverage with diverse descriptions.

Several works \cite{ic3, chatgptasks, socraticmodels} attempt to address this issue by applying a large language model (LLM) as a post-processing tool for guiding the model's output.
While a LLM enhances the quality of the generated captions, this approach incurs a substantial computational overhead.
The simultaneous execution of both a base model and the LLM during inference requires ample resources.

We observe a significant improvement in the performance of recently developed LLMs \cite{instructgpt, llama, llama2}.
This is achieved primarily by leveraging the three-stage training procedure described below.
The first two stages are common. The first stage involves pre-training on a large chunk of unsupervised data drawn from the Internet.
During the second stage, the model is fine-tuned using supervised data.
This data consists of human-labeled pairs of prompts and desired responses.
The key difference is in the third stage.
During this stage, a model is fine-tuned using reinforcement learning techniques~\cite{A2C, PPO}, maximizing the scores given by a reward model.
A reward model is trained given a prompt and several generated outputs.
A human labeler ranks the outputs from best to worst, and this ranking is used to train the reward model.
The question arises whether the well-established RL-based fine-tuning approach in NLP can be adapted to the image captioning task.

In this work, we present a novel approach of fine-tuning a pre-trained image captioning model with reinforcement learning, using a vision-language model as an off-the-shelf reward model.
It is similar to the methods described earlier but offers a distinct advantage due to its unsupervised nature.
Removing the need for supervised data or any kind of human labeling process greatly simplifies the task of data preparation and makes training high-quality image captioning models more affordable.

The key advantages of our method:
\begin{itemize}
    \item \textbf{Unsupervised.} Our method does not require any kind of human-labeled data during training.
    \item \textbf{No overhead.} During inference, the base model weights are simply replaced with the fine-tuned ones.
    \item \textbf{High detail captions.} Using BLIP2 \cite{BLIP2} as a baseline model, our method reaches remarkable 0.90 CLIP \cite{clipopenai} Recall (R@1) score on MS-COCO dataset \cite{mscoco} (Karpathy Test Split).
\end{itemize}

%% file: sect_related_work.tex
\subsection{Image captioning}
The use of alt-texts as an unsupervised data source has gained widespread adoption for pre-training vision-language models due to its effectiveness and scalability \cite{clipopenai, align, laion400m, laion5b}.
During training, one chooses to use contrastive \cite{clipopenai}, captioning \cite{cappa} or both \cite{coca} objectives to produce high-quality vision backbones.

A typical image captioning model comprises two main components: a vision encoder and a text decoder.
The vision encoder's features are shared with the text decoder using cross-attention \cite{coca, flamingo, BLIP} or projection to the decoder's embedding space, serving as a prefix prompt to guide the text generation process \cite{BLIP2}.

It is possible to train an entire model from scratch \cite{coca} or use a pre-trained backbone both for vision and text parts \cite{flamingo, BLIP, BLIP2}.
The most common choices are GPT-like text decoder and CLIP-trained image encoder.
\subsection{BLIP2}
BLIP2 is a vision-language model consisting of a frozen pre-trained image encoder, a LLM, and a trainable Q-Former between them. Q-Former is a lightweight Querying Transformer with Learned Queries (queries). 
The Q-Former is trained in two stages: Representation Learning and Generative Learning.
During both stages, Q-Formers' queries interact with image encoder output using a cross-attention mechanism, thus extracting all available visual information.

During Representation Learning, the Q-Former is trained jointly on three objectives: Image-Text Contrastive Learning (ITC), Image-Text Matching (ITM) and Image-grounded Text Generation (ITG). 
The ITC processes images and texts independently (like CLIP), whereas the ITM and the ITG require joint interaction of (image, text) pairs.

Let's have a closer look at the ITM since we use its outputs as a reward in our experiments.
Given a pair of (image, text), the output queries embeddings $Z$ (32 $\times$ 768) are fed into a two-class linear classifier (FC) to obtain logits, which are then averaged. 
This can be loosely written as follows:
\[ 
\text {ITM}(\text{image}, \text{text}) = \frac{1}{32}\sum_{i=1}^{32} \text{softmax}(\text{FC}(Z_i))[0]
\]

During Generative Learning, the Q-Former is connected to a frozen LLM via queries, which are projected to the LLM's feature space and act as a prefix, or so-called ``soft visual prompts''.
In case of a decoder-based LLM, which is used in our work, the Q-former is trained with the language modeling loss.

\subsection{Improving image captioning using additional LLM}
Recent works propose methods of applying LLMs to enhance the quality of captions generated by a base model. 

IC3 \cite{ic3} proposes to sample multiple (e.g. 10) captions for the given image using a base model with temperature-based sampling, obtaining representative samples from semantic distribution.
Then, a LLM aggregates sampled captions into a more detailed one using an engineered summary prompt.

``ChatGPT asks, BLIP2 answers'' \cite {chatgptasks} proposes to leverage the VQA capabilities of a base model.
When a base model generates an initial caption, a LLM starts asking questions about the image given the initial caption, thus gradually extracting more and more details that were not present in the initial caption.

Although both methods truly provide high detail captions, they share a common drawback.
These methods require multiple passes through the base model as well as a LLM running in parallel with the base model, which creates an excessive amount of computational resources required.





%% file: sect_method.tex
Our method is aimed at fine-tuning a captioning model in order to make generated captions more detailed with respect to the given image.
We use BLIP2 OPT-2.7B~\cite{BLIP2, zhang2022opt} as an image captioning model in our experiments.
The fine-tuning method is based on reinforcement learning, where a reward is calculated using similarity scores provided by vision-language models (e.g. CLIP, BLIP2-ITM) and a set of heuristics described further in this paper.
Note that our method does not introduce any new layers to a captioning model but only modifies the existing ones.

The training process is as follows.
At first, for a given image, a caption is generated by a captioning model.
Then, the reward is calculated for an (image, caption) pair using a vision-language model.
After that, the model's weights are updated using Advantage Author Critic (A2C) algorithm to generate a higher reward.

The intuition behind this algorithm is based on the empirical observation that more detailed descriptions typically yield higher vision-language similarity scores compared to those with fewer details.

This approach is similar to RLHF~\cite{RLHF}, the difference lies in the reward model.
In the case of RLHF, the reward model is trained on a special dataset that reflects human preferences.
In our approach, we don't train the reward model. Instead, we use a pre-trained vision-language model and a combination of heuristics.

\subsection{Metrics}
For the image captioning task, there are three main aspects of the caption quality that first come to mind:
\begin{enumerate}
  \item \textbf{Level of details.} The number of details that the model is able to describe: what is represented in the picture, colors of objects, etc.
  \item \textbf{Grammar.} The generated text must be grammatically correct (not a bag of words). 
  \item \textbf{Hallucinations.} Only the details present in the picture must be included in the text. 
\end{enumerate}

At the time of writing, there was no well-established benchmark that could be used to accurately assess the quality of the trained model in terms of the aspects mentioned above.
Therefore, in order to compare the models from different experiments, the authors manually examined the generated captions for random images to decide whether there was an improvement.

We also used the CLIP Recall metric on MS-COCO (Karpathy Test Split) for comparison with other works.
Firstly, captions for a set of images are generated using the tested captioning model.
Then, the text-to-image retrieval protocol for generated captions is run using the CLIP model.

The high value of CLIP Recall metric is correlated with the number of described objects and details in the image, however, the generated text may be grammatically incorrect or may contain hallucinations.
Moreover, our analysis shows that captions generated by a model with a high metric value may be less preferred by a human than captions generated by a weaker (in terms of the metric) model.

\subsection{Trainable parameters}

\begin{figure*}
	\centering
	
	\subcaptionbox{Step 1.
	Caption generation and $R(t_k)$ return computation.
	All parameters are frozen. Returns $R$ and tokens are passed to the next steps.
	  \label{fig:model:gen}}{\includegraphics[width=0.9\textwidth]{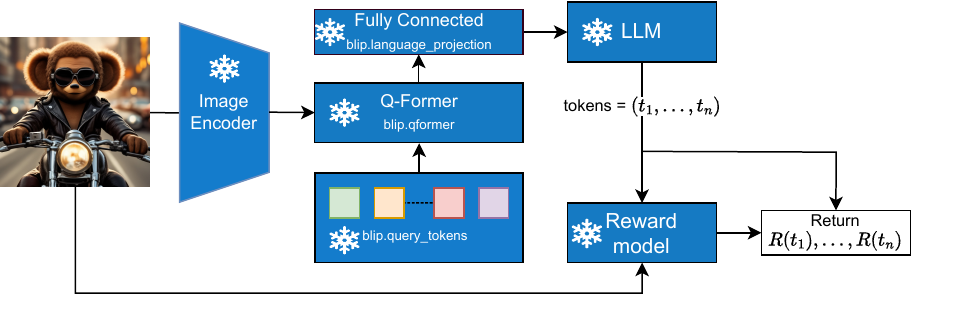}}

	\subcaptionbox{Step 2.
	Values computation and gradient update for the reward-prediction part.
	Values are computed for each token of a generated sequence.
	Trainable parametes are LLM's LoRa and value head.
	Values $V$ are passed to the third step.
	The architecture of the value head is shown in Figure~\ref{fig:vhead}.
	   \label{fig:model:value}}{\includegraphics[width=0.9\textwidth]{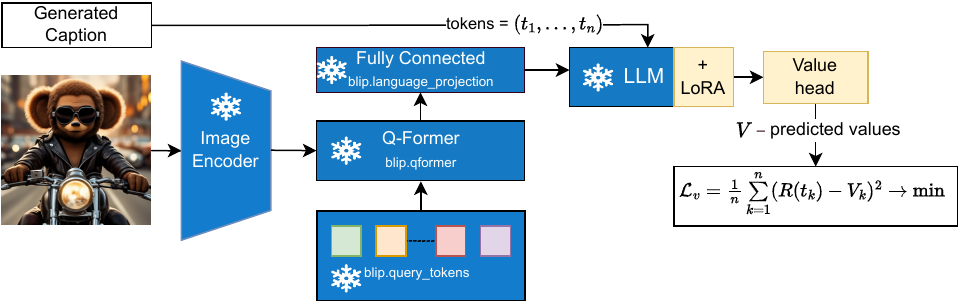}}
	
	\subcaptionbox{Step 3.
	Gradient update computation for the text generation part.
	The goal of the gradient update is to increase the probability of those tokens for which the value of the reward is higher than the average and decrease the probability otherwise.
	The operator $[\cdot]_{\text{sg}}$ means stop gradient or \texttt{.detach()} in PyTorch notation.
	  \label{fig:model:text}}{\includegraphics[width=0.9\textwidth]{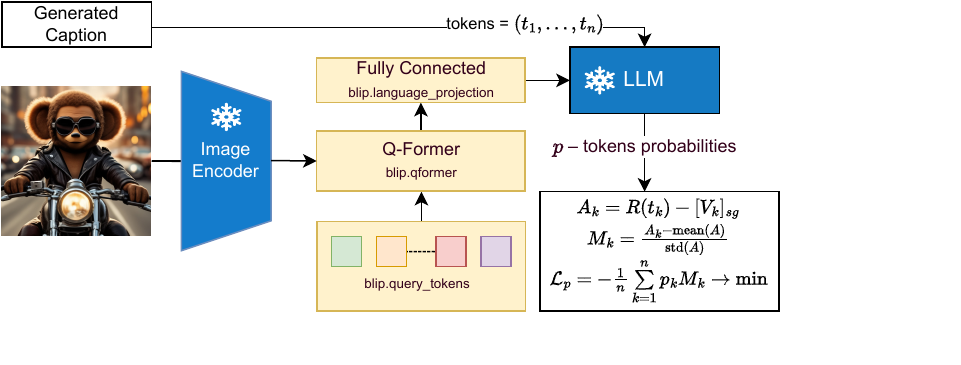}}

	\caption{The illustration of the training iteration.}
	\label{fig:model}
\end{figure*}

Trainable parameters are (see Figure~\ref{fig:model}): Q-Former (together with queries and language projection), value head, and LLM's LoRA parameters.
The image encoder and LLM are kept frozen. 
\begin{center}
\begin{tabular}{|l|l|}\hline
Part & \#Params \\\hline
Q-Former &  107M \\\hline
Value head & 275M \\\hline
LLM LoRA & 6M \\\hline
Total & 388M\\\hline
\end{tabular}
\end{center}

\begin{figure}
	\centering
	\includegraphics[width=3cm]{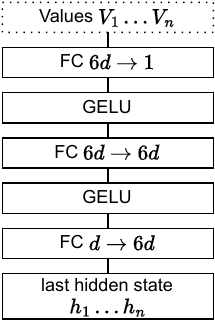}
	\caption{The architecture of the value head.}
	\label{fig:vhead}
\end{figure}

The architecture of the value head is shown in Figure~\ref{fig:vhead}.

\subsection{Dataset}

For training, we use keyframes from 9.5M randomly chosen YouTube videos.
We note that after training on 3M images (about 700 training steps), it becomes difficult to determine which checkpoint is better from a human perspective -- at 700 steps, or at 1700 steps, and so on.
At the same time, the value of the R@1 metric continues to increase on MS-COCO (Karpathy Test Split) until 5000 steps.


\subsection{Reward}\label{sect:reward}

\begin{table}
\begin{center}\begin{tabular}{|l|r|r|r|r|l|}
  token                & bad     & repeat & ref   & sim   & $R(t_k)$\\\hline
	{\color{red}an}      & 1       & 0      & $r_0$ & $r_1$ & $r_0 + r_1 - 7$ \\
	{\color{red}image}   & 1       & 0      & $r_0$ & $r_1$ & $r_0 + r_1 - 6$ \\
	{\color{red}of}      & 1       & 0      & $r_0$ & $r_1$ & $r_0 + r_1 - 5$ \\
	a                    & 0       & 0      & $r_0$ & $r_1$ & $r_0 + r_1 - 4$ \\
	man                  & 0       & 0      & $r_0$ & $r_1$ & $r_0 + r_1 - 4$ \\
	in                   & 0       & 0      & $r_0$ & $r_1$ & $r_0 + r_1 - 4$ \\
	green                & 0       & 0      & $r_0$ & $r_1$ & $r_0 + r_1 - 4$ \\
	{\color{orange}shirt}                & 0       & 0      & $r_0$ & $r_1$ & $r_0 + r_1 - 4$ \\
	and                  & 0       & 0      & $r_0$ & $r_1$ & $r_0 + r_1 - 4$ \\
	green                & 0       & 0      & $r_0$ & $r_1$ & $r_0 + r_1 - 4$ \\
	{\color{orange}shirt}   & 0       & 1      & $r_0$ & $r_1$ & $r_0 + r_1 - 4$ \\
	{\color{red}is}      & 1       & 0      & $r_0$ & $r_1$ & $r_0 + r_1 - 3$ \\
	{\color{red}talking} & 1       & 0      & $r_0$ & $r_1$ & $r_0 + r_1 - 2$ \\
	{\color{red}about}   & 1       & 0      & $r_0$ & $r_1$ & $r_0 + r_1 - 1$ \\
	a                    & 0       & 0      & $r_0$ & $r_1$ & $r_0 + r_1$ \\
	car                  & 0       & 0      & $r_0$ & $r_1$ & $r_0 + r_1$ \\
	eos                  & 0       &        &       &       &        
\end{tabular}\end{center}
  \caption{The example of computing the discounted return $R(t_k)$. The set of bad phrases is \{`an image of', `is talking about'\}. The repeated word is `shirt'. Note that the word `green' is not penalized since it represents color.}\label{tab:example-return}
\end{table}

Let us introduce the following notations.
For a given image, a model generates a caption consisting of tokens $(t_1,\ldots,t_n)$.
A text is a string consisting of decoded tokens.
The discounted return for a token $t_k$ is denoted as $R(t_k)$.
In our experiments, we use the discount factor $\gamma = 1$.

The discounted return $R(t_k)$ constists of five components:
\begin{enumerate}[topsep=8pt,itemsep=4pt,partopsep=4pt, parsep=4pt]
	\item sim(image, text)  = BLIP2-ITM(image, text)\\
	A similarity score, which is obtained from a vision-language model.
	This component rewards the model for generating a description with a lot of detail.
	In our case, the positive logit of the BLIP2-ITM model output (before softmax) is used as a similarity score.
	\item $\text{ref}(\text{text}) = \frac{1}{n} \sum_{k=1}^n \text{log}(p_k)$\\
	A logarithm of perplexity, which is computed with a reference model.
	This component rewards the model for the naturalness of the text.
	We use OPT-2.7b as a reference model. (see Section~\ref{sect:reference-component})
	\item $\text{bad}(s, t_1,\ldots,t_n) = \left\{\begin{array}{ll} 1, & {\exists} {s_1} {\le} {s} {\le} {s_2} \; t_{s_1}{,}{\ldots}{,}{t_{s_2}} {\in} \\ & \text{bad phrases} \\ 0, & \text{otherwise} \end{array}\right.$\\
	A penalty for the usage of prefixes (an image of, a youtube video of, \ldots), years (in the format of dddd), etc. (see Section~\ref{sect:badword-pen})
    \item $\text{repeat}(s, t_1,\ldots,t_n) = \left\{\begin{array}{ll} 1, & \text{if the word which} \\ & \text{contains } t_s \text{ already}\\ & \text{exists in the text} \\ 0, & \text{otherwise} \end{array}\right.$\\
	A penalty for repetitive words. As an exception, we don't penalize for repeated colors, prepositions, and articles.
	\item $\text{noeos}(t_1,\ldots,t_n) = \left\{\begin{array}{ll} 0, & \text{if } t_n = \text{eos} \\ 1,& \text{otherwise} \end{array}\right.$\\
	A penalty for not generating the end-of-sequence token.
\end{enumerate}

Overall, the return is computed as follows:
\begin{equation}
	\begin{array}{lcl}
		R(t_k) & = & \text{sim}(\text{text}, \text{image}) \\
		   & + & \text{ref}(\text{text}) \\
		   & - & \text{noeos}(t_1,\ldots,t_n) \\
		   & - & \sum_{s \ge k}\text{bad}(s, t_1,\ldots,t_n) \\
		   & - & \sum_{s \ge k}\text{repeat}(s, t_1,\ldots,t_n)
	\end{array}
	\label{eq:reward}
\end{equation}

The example of computing $R(t_k)$ is shown in Table~\ref{tab:example-return}.


\subsection{Training steps}

Each training iteration consists of three steps, shown in Figure~\ref{fig:model}.

\subsubsection {Step 1}
The first step, shown in Figure~\ref{fig:model:gen}, generates a caption, consisting of tokens $(t_1,\ldots,t_n)$, for a given image.
Then, for the (image, caption) pair, a similarity score is obtained from a vision-language model.
Finally, $R(t_k)$ is obtained using Equation~\ref{eq:reward}.
Returns $\left( R(t_1), \ldots, R(t_n) \right)$ and generated tokens are passed to the next steps. 

\subsubsection {Step 2}
The second step, shown in Figure~\ref{fig:model:value}, computes $V_k$ -- predictions for the expected discounted return $R(t_k)$, called `values'.
The task of this step is to compute values and perform a gradient step towards a more accurate prediction of $R(t_k)$:
\begin{equation}
  \mathcal{L}_v = \frac{1}{n}\sum_{k=1}^n (R(t_k) - V_k)^2 \rightarrow \text{min}
\end{equation}
To compute values in this step, we add trainable LoRA parameters to the LLM.
Note that these parameters participate only in the computation of the values and do not take part in the generation.
Thus, the part of the model responsible for the $R(t_k)$ predictions does not affect the generative part through the gradient step.
The difference $A_k = R(t_k) - V_k$, called `advantage', is passed to the third step.

\subsubsection {Step 3}
In the third step, shown in Figure~\ref{fig:model:text}, a gradient step is performed to increase the probability $p_k$ of generating a good token or decrease the probability of a bad token.
The notion of good and bad is based on $M_k$ -- a normalized value of $A_k$:
\begin{equation}
  M_k = \frac{A_k - \text{mean}(A)}{\text{std}(A)}\, ,
\end{equation}
where mean and std are taken across the whole batch, meaning that the tensor $A$ has the shape \texttt{(batch\_size, seq\_length)}.

The loss for a given caption then takes the form:
\begin{equation}\label{eq:gen-loss}
  \mathcal{L}_p = -\frac{1}{n}\sum_{k=1}^n p_k M_k \rightarrow \text{min}
\end{equation}

Intuitively, this means that if $A_k$ is strongly above average, the probability $p_k$ of a token $t_k$ increases, if it is strongly below average, it decreases, and if the value of $A_k$ is close to the average, the probability remains roughly unchanged.
Note that in the loss $\mathcal{L}_p$ (Equation~\ref{eq:gen-loss}) the gradient goes only to the probability $p_k$, while $M_k$ is used as a constant.

\subsection{Hyperparameters}

We use 20 warmup steps where the learning rate increases from 0 to 1e-5.
After that, the learning rate does not change (lr=1e-5).
We use the Adam~\cite{Adam} optimizer without weight decay.
The batch size is equal to 4096, meaning 4096 captions are generated within each step, where each generated caption is used exactly once in gradient computation.
We clamp the gradient with threshold 1.
During training, we use top\_k=6 sampling with temperature T=2.

%% file: sect_exp.tex
\subsection{Trainable parameters}\label{sect:exp-trainable-param}

Recall that in our approach, we fine-tune the following parts: query tokens, Q-Former, language projection.
Our experiments show that this set of parameters for fine-tuning provides the best result.
Other combinations were also tried (during Step 3, shown in Figure~\ref{fig:model:text}):

\begin{itemize}
    \item query tokens
    \item language projection
    \item query tokens, Q-Former, language projection, LoRA for the image encoder
    \item query tokens, Q-Former, language projection, LoRA for the LLM
\end{itemize}

However, other combinations lead to worse quality from the authors' point of view.

\subsection{Value head}

At first, one fully connected layer ($\text{hidden\_size} \rightarrow 1$) was used as a value head.
However, we discovered that such a value head is not sufficient for the given task: at first, the network trains as expected, but after a certain number of training steps, it collapses and starts generating incoherent text.
This problem was resolved after the adoption of a more complex architecture for the value head, shown in Figure~\ref{fig:vhead}.

Note that the architecture proposed in Figure~\ref{fig:vhead} has a significantly larger number of parameters compared to a commonly used single fully connected layer architecture (275M vs. 2560).

\subsection{Reference component}\label{sect:reference-component}

This component is necessary if we want to generate human-readable captions.
Without this component, the model will generate bizarre, unnatural captions (from a human perspective).
At the same time, similarity scores and, as a sequence, CLIP Recall metrics will be high, as shown in Tables~\ref{tab:clip-recall},\,\ref{tab:color-cmp},\,\ref{tab:coco-captions}, where VLRM was trained with a reference component, and VLRM-RS was trained without it.

Our experiments show that if we use OPT-2.7B (from BLIP2) as a reference model, the generated captions look more natural.
We also tried to use LLaMa-2-7B~\cite{llama2} as a reference model, but in that case, captions looked worse than the baseline OPT-2.7B model.

\subsection{Bad phrases penalty}\label{sect:badword-pen}

During training, the model finds meaningless prefixes that increase the score provided by a vision-language model:
`an image of', `a video of',  `a camera shot of', etc.
As a result, the model starts each caption with one or more such prefixes.
Usually, they are out of place and do not provide any useful information.

The model also tends to specify the year when the image was taken in the generated caption (e.g., ``in 1993''), even though it is obviously impossible to determine in which year the action took place.

If the image shows a person during conversation, the model tends to generate ``a person \textit{is talking about $<$smth$>$}'',
which is clearly a hallucination, since the model has no access to audio, and, as a consequence, cannot determine the subject of conversation.

To eliminate the unwanted generation tendencies described above, a bad phrase penalty is introduced (see Section~\ref{sect:reward}).
This penalty is assigned to a token if it is contained in a bad phrase.
In Table~\ref{tab:badphr}, we list the bad phrases that we have discovered throughout the experiments.

%% file: sect_results.tex
In this section, we present two models obtained by our method.
The first one, called VLRM, is proposed as the main result.
This model generates texts similar in style to those generated by the original BLIP2, but with more details.
The second proposed model, called VLRM-RS (Retrieval Specialization), is trained with the aim of obtaining the highest CLIP Recall value.

The difference between VLRM and VLRM-RS lies in the computation of $R(t_k)$.
Firstly, VLRM-RS is trained without a $\text{ref}(\text{text})$ component.
Secondly, a CLIP model-based summand is added to the $\text{sim}$ component as shown in Equation~\ref{eq:oursRS-sim}.
We use OpenClip~\cite{openclip} CLIP-ViT-bigG-14-laion2B-39B-b160k as a CLIP model.

\begin{equation}
\begin{array}{c}
  S_{ij} = \text{ cos}(\text{clip}(\text{image}_i), \text{clip}(\text{text}_j)) \\
  D = (\text{softmax}(S,0) * \text{softmax}(S,1)).\text{diag} \\ 
  \Phi = \frac{D - \text{mean}(D)}{\text{std}(D)} \\
  \text{sim}( \text{image}_k, \text{text}_k) = \text{blip2\_itm}(\text{image}_k, \text{text}_k) + 0.3 \Phi_k
\end{array}\label{eq:oursRS-sim}
\end{equation}

On MS-COCO Karpathy Test Split (see Table~\ref{tab:clip-recall}), VLRM and VLRM-RS show +38.8\% and +41.5\% gain in R@1 compared to original BLIP2, respectively.
For CLIP Recall computing, we center crop images both during captioning and retrieval.
Here, we use OpenClip~\cite{openclip} CLIP-ViT-g-14-laion2B-s12B-b42K as a CLIP model, same as in IC3~\cite{ic3}.

In Table~\ref{tab:color-cmp}, we compare the original BLIP2 and our models in terms of color coverage. 
In these examples, we can see that our modifications incorporate colors much more actively than the original model.
Also note that if an object has multiple colors, the model has the property to list them with commas, see the 4th image in Table~\ref{tab:coco-captions}.

In Table~\ref{tab:infer-params}, we list the set of generation parameters used in all cases during inference.
From Tables~\ref{tab:clip-recall},\,\ref{tab:color-cmp},\,\ref{tab:coco-captions} we can see that our models typically generate much longer and more detailed descriptions.

\begin{table}
        \begin{center}
        \begin{tabular}{l|c}
        \toprule
        min\_new\_tokens& 4\\ \hline
        max\_new\_tokens& 60\\ \hline
        do\_sample& false\\ \hline
        no\_repeat\_ngram\_size& 2\\ \hline
        num\_beams& 5\\ 
        \bottomrule
        \end{tabular}
        \end{center}
          \caption{Inference generation parameters.}\label{tab:infer-params}
        \end{table}
\begin{table}
\centering
\begin{tabular}{l*{4}{c}}

    \toprule
    MODEL       & MRR $\uparrow$ & R@1 $\uparrow$ & R@5$\uparrow$ & R@10 $\uparrow$ \\
    \midrule
    REF         & 0.583	& 0.471	& 0.712	& 0.801 \\
    BLIP2       & 0.625	& 0.517	& 0.748	& 0.832 \\
    \midrule
    BLIP2 + $\text{IC}^3$\cite{ic3} & 0.746 & 0.652 & 0.863 & 0.921 \\
    \midrule
    VLRM        & 0.940	& 0.905	& 0.984	& 0.994 \\ 
    VLRM-RS     & 0.958	& 0.932	& 0.991	& 0.997 \\ 
    \bottomrule
\end{tabular}
\vspace{2mm}
    \caption{CLIP Recall for generated captions in the MS-COCO Dataset (Karpathy Test Split). MRR: Mean Reciprocal Recall, R@K: Recall @ K.
             REF means ground-truth human-labeled captions. BLIP2 means original model. 
             }
\label{tab:clip-recall}
\end{table}

\newlength{\colMethodWidth}
\setlength{\colMethodWidth}{\maxof{\widthof{VLRM-RS}}{\widthof{Method}}}
\newlength{\colScoreWidth}
\setlength{\colScoreWidth}{\widthof{-0.000}}
\begin{table*}
    \begin{tabular}{ll}
        &
        \begin{tabular}{p{\colMethodWidth}p{\colScoreWidth}p{\colScoreWidth}l}
            Method  & ITM    & CLIP  & \parbox{0.4\textwidth}{Generated Caption} \\
        \end{tabular}\\
        
        &\\
        \hline
        &\\
        \raisebox{-.5\height}{\includegraphics[width=5cm]{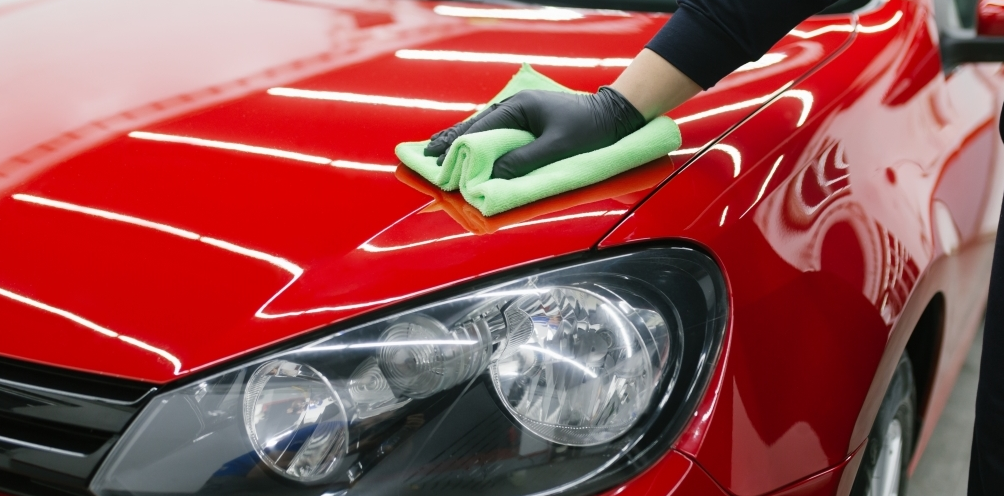}}
        &
        \begin{tabular}{p{\colMethodWidth}p{\colScoreWidth}p{\colScoreWidth}l}
            BLIP2 & 3.148 & 0.326  & \parbox{0.4\textwidth}{a person cleaning the hood of a \R{red} car} \\
                  &&& \\
            VLRM   & 4.881 & 0.342 & \parbox{0.4\textwidth}{a person in a \R{black} shirt with \R{black} gloves is using a \R{green} towel to clean the forehead of the hood of a \R{red} volkswagen sedan} \\
                  &&& \\
            VLRM-RS   & 5.111 & 0.343 & \parbox{0.4\textwidth}{hand with \R{black} gloves applying \R{green} blob towel onto \R{red} Volkswagen Golf headlight hood with shine on \R{red} vehicle maintenance workshop web stock} \\
        \end{tabular}\\
        
        &\\
        \hline
        &\\

        \raisebox{-.5\height}{\includegraphics[width=5cm]{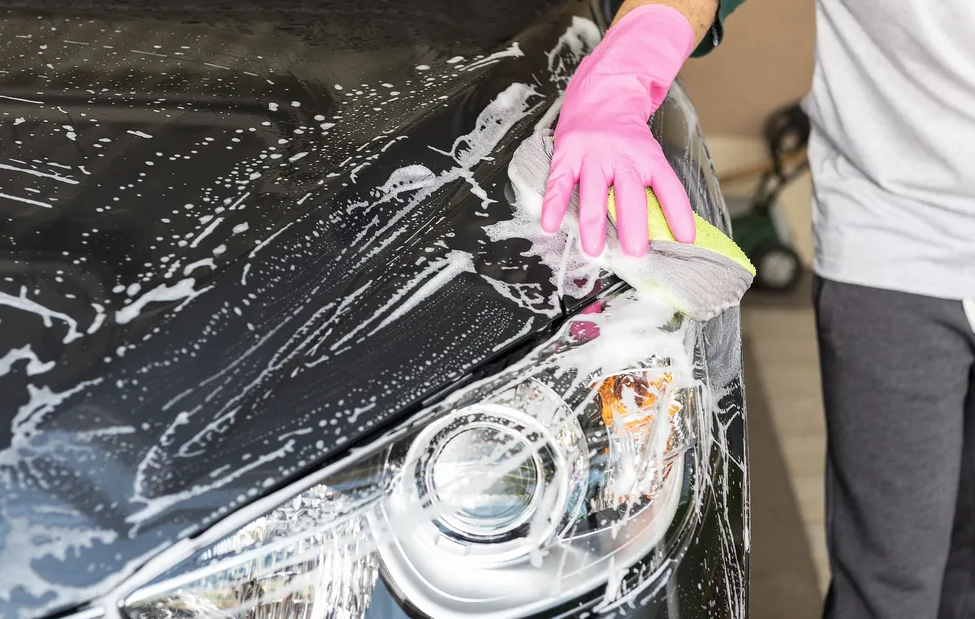}}
        &
        \begin{tabular}{p{\colMethodWidth}p{\colScoreWidth}p{\colScoreWidth}l}
            BLIP2   & 2.232 & 0.319 & \parbox{0.4\textwidth}{a person is washing a car with a sponge} \\
                    &&& \\
            VLRM    & 4.543 & 0.318 & \parbox{0.4\textwidth}{an asian man in a \R{white} sweatshirt with \R{pink} gloves cleaning the headlight of the hood of a \R{black} sedan with foamy soap in the garage} \\
                    &&& \\
            VLRM-RS & 4.824 & 0.333 & \parbox{0.4\textwidth}{employee with \R{pink} gloves \R{white} shirt washing black sedan headlight with foam and bubbles washing in garage automotive services NSW stock photo} \\
        \end{tabular} \\
        
        &\\
        \hline
        &\\

        \raisebox{-.5\height}{\includegraphics[width=5cm]{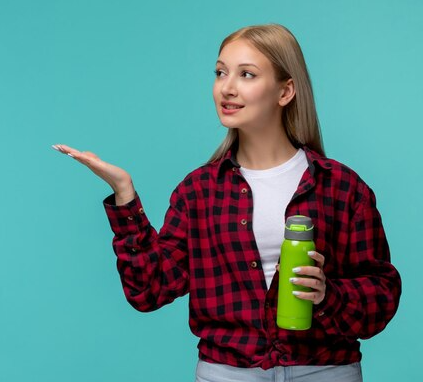}}
        &
        \begin{tabular}{p{\colMethodWidth}p{\colScoreWidth}p{\colScoreWidth}l}
            BLIP2   & 4.031 & 0.364 & \parbox{0.4\textwidth}{young woman holding a water bottle and pointing to the side} \\
                    &&& \\
            VLRM    & 6.110 & 0.414 & \parbox{0.4\textwidth}{a girl with long \R{blonde} hair in a \R{red} and \R{black} checkered shirt gesturing with a \R{green} bottle and pointing it towards herself in front of a \R{blue} background} \\
                    &&& \\
            VLRM-RS & 6.163 & 0.435 & \parbox{0.4\textwidth}{a \R{blonde} teenage girl in \R{burgundy} plaid shirt speaking gestures with \R{green} bottle on \R{blue} backdrop} \\
        \end{tabular} \\

        &\\
        \hline
        &\\
        
        \raisebox{-.5\height}{\includegraphics[width=5cm]{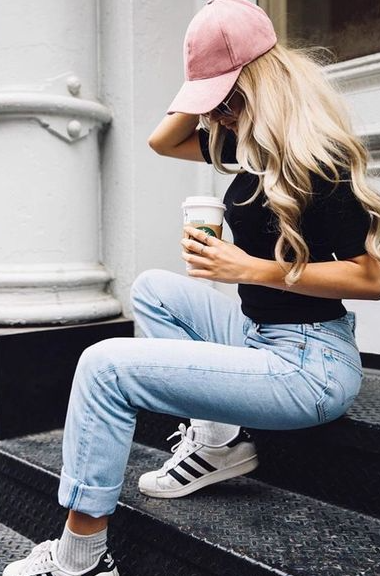}}
        &
        \begin{tabular}{p{\colMethodWidth}p{\colScoreWidth}p{\colScoreWidth}l}
            BLIP2   & 5.529 & 0.271 & \parbox{0.4\textwidth}{a \R{blonde} woman in jeans and a \R{pink} hat sitting on steps} \\
                    &&& \\
            VLRM    & 5.840 & 0.318 & \parbox{0.4\textwidth}{a girl with long \R{blonde} hair in a \R{black} t-shirt, \R{blue} jeans, \R{adidas} sneakers and a \R{pink} baseball cap sitting on stairs} \\
                    &&& \\
            VLRM-RS & 6.074 & 0.308 & \parbox{0.4\textwidth}{\R{blonde} woman in \R{black} shirt with \R{pink} cap and jeans drinking coffee on staircase} \\
        \end{tabular} \\

        &\\
        \hline

    \end{tabular}

    \caption{Comparison of the original BLIP2 and our models in terms of color usage.}\label{tab:color-cmp}

\end{table*}

\begin{table*}
    \begin{tabular}{ll}
        &
        \begin{tabular}{p{\colMethodWidth}p{\colScoreWidth}p{\colScoreWidth}l}
            Method  & ITM    & CLIP  & \parbox{0.4\textwidth}{Generated Caption} \\
        \end{tabular}\\
        
        &\\
        \hline
        &\\

        \raisebox{-.5\height}{\includegraphics[width=5cm]{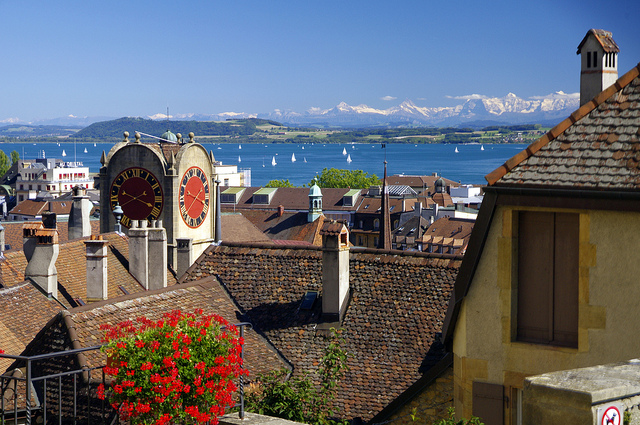}}
        &
        \begin{tabular}{p{\colMethodWidth}p{\colScoreWidth}p{\colScoreWidth}l}
            BLIP2   & -0.009 & 0.114 & \parbox{0.4\textwidth}{a clear blue sky} \\
                    &&& \\
            VLRM    & 3.152  & 0.240 & \parbox{0.4\textwidth}{roofs of a building with a red clock and a bunch of red flowers on it} \\
                    &&& \\
            VLRM-RS & 4.083  & 0.269 & \parbox{0.4\textwidth}{clear blue sky above roofs with red flowers and clock} \\
        \end{tabular}\\
        
        &\\
        \hline
        &\\

        \raisebox{-.5\height}{\includegraphics[width=5cm]{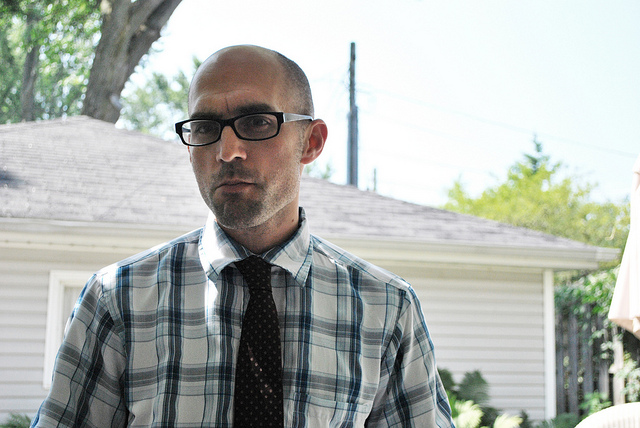}}
        &
        \begin{tabular}{p{\colMethodWidth}p{\colScoreWidth}p{\colScoreWidth}l}
            BLIP2   & 0.936 & 0.302 & \parbox{0.4\textwidth}{a man in a plaid shirt} \\
                    &&& \\
            VLRM    & 5.402 & 0.369 & \parbox{0.4\textwidth}{a bald man with glasses wearing a blue and white checkered shirt and a black tie standing in front of a house with a tree in the background} \\
                    &&& \\
            VLRM-RS & 5.549 & 0.399 & \parbox{0.4\textwidth}{a bald man with glasses wearing blue and white plaid shirt with black tie standing looking away from backyard} \\
        \end{tabular}\\
        
        &\\
        \hline
        &\\

        \raisebox{-.5\height}{\includegraphics[width=5cm]{imgs/tab_coco_captions/COCO_val2014_000000060069.jpg}}
        &
        \begin{tabular}{p{\colMethodWidth}p{\colScoreWidth}p{\colScoreWidth}l}
            BLIP2   & 1.565  & 0.233 & \parbox{0.4\textwidth}{a person is parasailing in the ocean} \\
                    &&& \\
            VLRM    & 3.552 &  0.333 & \parbox{0.4\textwidth}{a guy riding a kite with a black, white, and green design on it riding over a wave in the ocean} \\
                    &&& \\
            VLRM-RS & 4.010  & 0.312 & \parbox{0.4\textwidth}{a kite board with white and black design flying over a wave with a person on it} \\
        \end{tabular}\\

        &\\
        \hline
        &\\

        \raisebox{-.5\height}{\includegraphics[width=5cm]{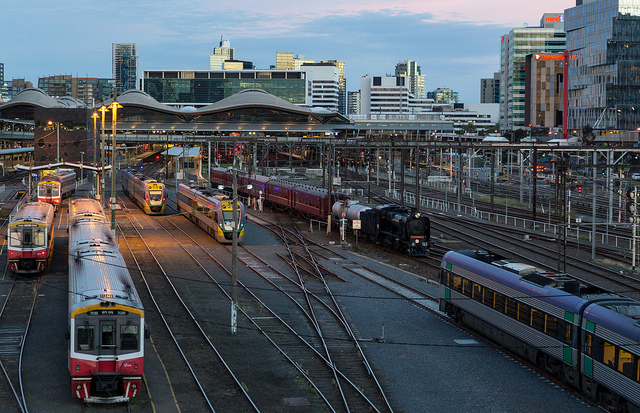}}
        &
        \begin{tabular}{p{\colMethodWidth}p{\colScoreWidth}p{\colScoreWidth}l}
            BLIP2   & 1.569 & 0.305 & \parbox{0.4\textwidth}{a train station with several trains on the tracks} \\
                    &&& \\
            VLRM    & 4.451 & 0.314 & \parbox{0.4\textwidth}{a bunch of red, grey, white, yellow, and black trains lined up on the tracks at dusk in a station with skyscrapers in the background} \\
                    &&& \\
            VLRM-RS & 4.153 & 0.302 & \parbox{0.4\textwidth}{a lot of red, grey and silver trains parked on tracks with dusk skyline city station} \\
        \end{tabular}\\
  
        &\\
        \hline
        &\\

        \raisebox{-.5\height}{\includegraphics[width=5cm]{imgs/tab_coco_captions/COCO_val2014_000000127955.jpg}}
        &
        \begin{tabular}{p{\colMethodWidth}p{\colScoreWidth}p{\colScoreWidth}l}
            BLIP2   & 2.690 & 0.330 & \parbox{0.4\textwidth}{two bears fighting in the water} \\
                    &&& \\
            VLRM    & 4.303 & 0.345 & \parbox{0.4\textwidth}{a pair of brown bears fighting in the water with their mouths open in front of a rock} \\
                    &&& \\
            VLRM-RS & 3.723 & 0.353 & \parbox{0.4\textwidth}{two brown bears fighting yelling in water} \\
        \end{tabular}\\
        
        &\\
        \hline

      \end{tabular}
      \caption{Examples of generation by the original BLIP2 and our models on random images from the MS-COCO dataset.}\label{tab:coco-captions}
\end{table*}

%% file: sect_conclusion.tex
We propose VLRM, a method for fine-tuning an existing image captioning model using reinforcement learning and vision-language models as reward models.
The method manages to significantly improve the generation quality without human-labeled data and is applicable to any image captioning model.
We consider VLRM as an important step towards building human-level multimodal AI systems.

%% file: sect_appendix.tex
\begin{table*}
\begin{center}
    \begin{tabular}{|*{5}{c|}}

    jf        & cam grabs    & footageshows    & tv camera shows     & camera shot footage           \\
    jpj       & instagram    & blurry scene    & camshot showing     & a zoomed in view of           \\
    clip      & on camera    & tv film show    & capture depicts     & in webcam is shown,           \\
    a tv      & that says    & screen photo    & televised image     & a zoomed in picture           \\
    zoom      & grey gray    & in a web cam    & images captures     & a blurry picture of           \\
    shot      & the tv ad    & screen image    & camera tv shows     & television ad image           \\
    in tv     & animation    & in a podcast    & tv view showing     & television displays           \\
    video     & there are    & channel view    & channel showing     & a zoomed-in view of           \\
    blurry    & a film of    & camera shows    & webcam captures     & a spherical view of           \\
    on-cam    & featuring    & screenscreen    & tv program shows    & google channel shows          \\
    cam of    & the camera   & this tv view    & a spherical view    & camera is talking to          \\
    in web    & tv picture   & on the right    & camera shotshows    & the television image          \\
    in cam    & film shows   & cam captures    & a blurry picture    & camera is looking at          \\
    tv view   & this is an   & in instagram    & a screenshare of    & middle of the screen          \\
    twitter   & screenshot   & tv picture of   & a zoomed in view    & a zoomed in image of          \\
    camera,   & in twitter   & film captures   & zoomed in screen    & a screen capture from         \\
    on- cam   & game shows   & photo showing   & tv image showing    & television image of a         \\
    webcast   & shot image   & channel shows   & cam footage show    & the camera is looking         \\
    in zoom   & tv view of   & on the screen   & a blurry blur of    & the camera is talking         \\
    footage   & text shows   & a close-up of   & in the middle of    & television view shows         \\
    youtube   & tv image of  & tv film shows   & cam footageshows    & a zoomed in picture of        \\
    camshot   & tv image is  & a webcam of a   & an image is shown   & a view from the camera        \\
    , image   & watch video  & a screen grab   & a blurry image of   & tv channel camera show        \\
    in chat   & in facebook  & tv view shows   & tv channel camera   & computer display shows        \\
    in msnbc  & camera show  & camera's view   & camera is looking   & television image shows        \\
    a blurry  & cam footage  & talking about   & screen image shot   & google channel showing        \\
    animated  & a zoomed in  & a close up of   & cam footage shows   & television view showing       \\
    in a cam  & tv ad shows  & tv ad showing   & screenscreen view   & tv channel camera shows       \\
    download  & a stream of  & camera tv show  & screen shot image   & the camera is talking to      \\
    photo of  & a screen of  & tv camera shot  & camera is showing   & the camera is looking at      \\
    close-up  & an image of  & tv image shows  & captured on video   & computer display showing      \\
    cam show  & in animated  & spherical view  & camera is talking   & television image showing      \\
    there is  & close up of  & on the web cam  & a zoomed in image   & game showcamera displays      \\
    tv shows  & on the left  & screen showing  & screen is showing   & multiple images captures      \\
    facebook  & cam showing  & camera footage  & television screen   & a camera taking a look at     \\
    close up  & tv ad image  & webcam footage  & camera shot shows   & zoomed-in picturezoomed-in    \\
    camera tv & the tv view  & a blurry image  & television showing  & youtube browser displaying    \\
    this is a & camera shot  & camera showing  & in webcam is shown  & television image showing a    \\
    cam shows & movie scene  & tv film showed  & vintage film shows  & in the middle of the screen   \\
    gray grey & camera view  & zoomed-in view  & television view of  & the camera focuses ona zoomed \\
    tv showed & screen shot  & a close-up of a & cameras angle shot  &                               \\
    tv screen & the cam has  & - the cam has a & television image of &                               \\
    a view of & screen print & blurry scene of & this tv image shows &                               \\

\end{tabular}
\end{center}
    \caption{The list of detected bad phrases.}\label{tab:badphr}
\end{table*}


%% file: references.bib
@misc{BLIP2,
      title={BLIP-2: Bootstrapping Language-Image Pre-training with Frozen Image Encoders and Large Language Models}, 
      author={Junnan Li and Dongxu Li and Silvio Savarese and Steven Hoi},
      year={2023},
      eprint={2301.12597},
      archivePrefix={arXiv},
      primaryClass={cs.CV}
}

@misc{BLIP,
      title={BLIP: Bootstrapping Language-Image Pre-training for Unified Vision-Language Understanding and Generation}, 
      author={Junnan Li and Dongxu Li and Caiming Xiong and Steven Hoi},
      year={2022},
      eprint={2201.12086},
      archivePrefix={arXiv},
      primaryClass={cs.CV}
}

@misc{chatgptasks,
      title={ChatGPT Asks, BLIP-2 Answers: Automatic Questioning Towards Enriched Visual Descriptions}, 
      author={Deyao Zhu and Jun Chen and Kilichbek Haydarov and Xiaoqian Shen and Wenxuan Zhang and Mohamed Elhoseiny},
      year={2023},
      eprint={2303.06594},
      archivePrefix={arXiv},
      primaryClass={cs.CV}
}

@misc{zhang2022opt,
      title={OPT: Open Pre-trained Transformer Language Models}, 
      author={Susan Zhang and Stephen Roller and Naman Goyal and Mikel Artetxe and Moya Chen and Shuohui Chen and Christopher Dewan and Mona Diab and Xian Li and Xi Victoria Lin and Todor Mihaylov and Myle Ott and Sam Shleifer and Kurt Shuster and Daniel Simig and Punit Singh Koura and Anjali Sridhar and Tianlu Wang and Luke Zettlemoyer},
      year={2022},
      eprint={2205.01068},
      archivePrefix={arXiv},
      primaryClass={cs.CL}
}

@misc{RLHF,
      title={Deep reinforcement learning from human preferences}, 
      author={Paul Christiano and Jan Leike and Tom B. Brown and Miljan Martic and Shane Legg and Dario Amodei},
      year={2023},
      eprint={1706.03741},
      archivePrefix={arXiv},
      primaryClass={stat.ML}
}

@misc{Adam,
      title={Adam: A Method for Stochastic Optimization}, 
      author={Diederik P. Kingma and Jimmy Ba},
      year={2017},
      eprint={1412.6980},
      archivePrefix={arXiv},
      primaryClass={cs.LG}
}

@misc{ic3,
      title={IC3: Image Captioning by Committee Consensus}, 
      author={David M. Chan and Austin Myers and Sudheendra Vijayanarasimhan and David A. Ross and John Canny},
      year={2023},
      eprint={2302.01328},
      archivePrefix={arXiv},
      primaryClass={cs.CV}
}

@misc{laion400m,
      title={LAION-400M: Open Dataset of CLIP-Filtered 400 Million Image-Text Pairs}, 
      author={Christoph Schuhmann and Richard Vencu and Romain Beaumont and Robert Kaczmarczyk and Clayton Mullis and Aarush Katta and Theo Coombes and Jenia Jitsev and Aran Komatsuzaki},
      year={2021},
      eprint={2111.02114},
      archivePrefix={arXiv},
      primaryClass={cs.CV}
}

@misc{laion5b,
      title={LAION-5B: An open large-scale dataset for training next generation image-text models}, 
      author={Christoph Schuhmann and Romain Beaumont and Richard Vencu and Cade Gordon and Ross Wightman and Mehdi Cherti and Theo Coombes and Aarush Katta and Clayton Mullis and Mitchell Wortsman and Patrick Schramowski and Srivatsa Kundurthy and Katherine Crowson and Ludwig Schmidt and Robert Kaczmarczyk and Jenia Jitsev},
      year={2022},
      eprint={2210.08402},
      archivePrefix={arXiv},
      primaryClass={cs.CV}
}

@misc{mscoco,
      title={Microsoft COCO: Common Objects in Context}, 
      author={Tsung-Yi Lin and Michael Maire and Serge Belongie and Lubomir Bourdev and Ross Girshick and James Hays and Pietro Perona and Deva Ramanan and C. Lawrence Zitnick and Piotr Dollár},
      year={2015},
      eprint={1405.0312},
      archivePrefix={arXiv},
      primaryClass={cs.CV}
}

@misc{conceptualcaptions,
      title={Conceptual 12M: Pushing Web-Scale Image-Text Pre-Training To Recognize Long-Tail Visual Concepts}, 
      author={Soravit Changpinyo and Piyush Sharma and Nan Ding and Radu Soricut},
      year={2021},
      eprint={2102.08981},
      archivePrefix={arXiv},
      primaryClass={cs.CV}
}

@misc{socraticmodels,
      title={Socratic Models: Composing Zero-Shot Multimodal Reasoning with Language}, 
      author={Andy Zeng and Maria Attarian and Brian Ichter and Krzysztof Choromanski and Adrian Wong and Stefan Welker and Federico Tombari and Aveek Purohit and Michael Ryoo and Vikas Sindhwani and Johnny Lee and Vincent Vanhoucke and Pete Florence},
      year={2022},
      eprint={2204.00598},
      archivePrefix={arXiv},
      primaryClass={cs.CV}
}

@misc{coca,
      title={CoCa: Contrastive Captioners are Image-Text Foundation Models}, 
      author={Jiahui Yu and Zirui Wang and Vijay Vasudevan and Legg Yeung and Mojtaba Seyedhosseini and Yonghui Wu},
      year={2022},
      eprint={2205.01917},
      archivePrefix={arXiv},
      primaryClass={cs.CV}
}

@misc{instructgpt,
      title={Training language models to follow instructions with human feedback}, 
      author={Long Ouyang and Jeff Wu and Xu Jiang and Diogo Almeida and Carroll L. Wainwright and Pamela Mishkin and Chong Zhang and Sandhini Agarwal and Katarina Slama and Alex Ray and John Schulman and Jacob Hilton and Fraser Kelton and Luke Miller and Maddie Simens and Amanda Askell and Peter Welinder and Paul Christiano and Jan Leike and Ryan Lowe},
      year={2022},
      eprint={2203.02155},
      archivePrefix={arXiv},
      primaryClass={cs.CL}
}

@misc{llama,
      title={LLaMA: Open and Efficient Foundation Language Models}, 
      author={Hugo Touvron and Thibaut Lavril and Gautier Izacard and Xavier Martinet and Marie-Anne Lachaux and Timothée Lacroix and Baptiste Rozière and Naman Goyal and Eric Hambro and Faisal Azhar and Aurelien Rodriguez and Armand Joulin and Edouard Grave and Guillaume Lample},
      year={2023},
      eprint={2302.13971},
      archivePrefix={arXiv},
      primaryClass={cs.CL}
}

@misc{llama2,
      title={Llama 2: Open Foundation and Fine-Tuned Chat Models}, 
      author={Hugo Touvron and Louis Martin and Kevin Stone and Peter Albert and Amjad Almahairi and Yasmine Babaei and Nikolay Bashlykov and Soumya Batra and Prajjwal Bhargava and Shruti Bhosale and Dan Bikel and Lukas Blecher and Cristian Canton Ferrer and Moya Chen and Guillem Cucurull and David Esiobu and Jude Fernandes and Jeremy Fu and Wenyin Fu and Brian Fuller and Cynthia Gao and Vedanuj Goswami and Naman Goyal and Anthony Hartshorn and Saghar Hosseini and Rui Hou and Hakan Inan and Marcin Kardas and Viktor Kerkez and Madian Khabsa and Isabel Kloumann and Artem Korenev and Punit Singh Koura and Marie-Anne Lachaux and Thibaut Lavril and Jenya Lee and Diana Liskovich and Yinghai Lu and Yuning Mao and Xavier Martinet and Todor Mihaylov and Pushkar Mishra and Igor Molybog and Yixin Nie and Andrew Poulton and Jeremy Reizenstein and Rashi Rungta and Kalyan Saladi and Alan Schelten and Ruan Silva and Eric Michael Smith and Ranjan Subramanian and Xiaoqing Ellen Tan and Binh Tang and Ross Taylor and Adina Williams and Jian Xiang Kuan and Puxin Xu and Zheng Yan and Iliyan Zarov and Yuchen Zhang and Angela Fan and Melanie Kambadur and Sharan Narang and Aurelien Rodriguez and Robert Stojnic and Sergey Edunov and Thomas Scialom},
      year={2023},
      eprint={2307.09288},
      archivePrefix={arXiv},
      primaryClass={cs.CL}
}

@misc{clipopenai,
      title={Learning Transferable Visual Models From Natural Language Supervision}, 
      author={Alec Radford and Jong Wook Kim and Chris Hallacy and Aditya Ramesh and Gabriel Goh and Sandhini Agarwal and Girish Sastry and Amanda Askell and Pamela Mishkin and Jack Clark and Gretchen Krueger and Ilya Sutskever},
      year={2021},
      eprint={2103.00020},
      archivePrefix={arXiv},
      primaryClass={cs.CV}
}

@software{openclip,
  author       = {Ilharco, Gabriel and
                  Wortsman, Mitchell and
                  Wightman, Ross and
                  Gordon, Cade and
                  Carlini, Nicholas and
                  Taori, Rohan and
                  Dave, Achal and
                  Shankar, Vaishaal and
                  Namkoong, Hongseok and
                  Miller, John and
                  Hajishirzi, Hannaneh and
                  Farhadi, Ali and
                  Schmidt, Ludwig},
  title        = {OpenCLIP},
  month        = jul,
  year         = 2021,
  note         = {If you use this software, please cite it as below.},
  publisher    = {Zenodo},
  version      = {0.1},
  doi          = {10.5281/zenodo.5143773},
  url          = {https://doi.org/10.5281/zenodo.5143773}
}

@misc{PPO,
      title={Proximal Policy Optimization Algorithms}, 
      author={John Schulman and Filip Wolski and Prafulla Dhariwal and Alec Radford and Oleg Klimov},
      year={2017},
      eprint={1707.06347},
      archivePrefix={arXiv},
      primaryClass={cs.LG}
}

@misc{flamingo,
      title={Flamingo: a Visual Language Model for Few-Shot Learning}, 
      author={Jean-Baptiste Alayrac and Jeff Donahue and Pauline Luc and Antoine Miech and Iain Barr and Yana Hasson and Karel Lenc and Arthur Mensch and Katie Millican and Malcolm Reynolds and Roman Ring and Eliza Rutherford and Serkan Cabi and Tengda Han and Zhitao Gong and Sina Samangooei and Marianne Monteiro and Jacob Menick and Sebastian Borgeaud and Andrew Brock and Aida Nematzadeh and Sahand Sharifzadeh and Mikolaj Binkowski and Ricardo Barreira and Oriol Vinyals and Andrew Zisserman and Karen Simonyan},
      year={2022},
      eprint={2204.14198},
      archivePrefix={arXiv},
      primaryClass={cs.CV}
}

@misc{cappa,
      title={Image Captioners Are Scalable Vision Learners Too}, 
      author={Michael Tschannen and Manoj Kumar and Andreas Steiner and Xiaohua Zhai and Neil Houlsby and Lucas Beyer},
      year={2023},
      eprint={2306.07915},
      archivePrefix={arXiv},
      primaryClass={cs.CV}
}

@misc{align,
      title={Scaling Up Visual and Vision-Language Representation Learning With Noisy Text Supervision}, 
      author={Chao Jia and Yinfei Yang and Ye Xia and Yi-Ting Chen and Zarana Parekh and Hieu Pham and Quoc V. Le and Yunhsuan Sung and Zhen Li and Tom Duerig},
      year={2021},
      eprint={2102.05918},
      archivePrefix={arXiv},
      primaryClass={cs.CV}
}

@misc{A2C,
      title={Asynchronous Methods for Deep Reinforcement Learning}, 
      author={Volodymyr Mnih and Adrià Puigdomènech Badia and Mehdi Mirza and Alex Graves and Timothy P. Lillicrap and Tim Harley and David Silver and Koray Kavukcuoglu},
      year={2016},
      eprint={1602.01783},
      archivePrefix={arXiv},
      primaryClass={cs.LG}
}
